\documentclass{article}

\usepackage[preprint]{neurips_2024}

\usepackage{subcaption}
\usepackage{booktabs}
\usepackage{pifont}
\newcommand{\cmark}{\ding{51}}%
\newcommand{\xmark}{\ding{55}}%
\usepackage{enumitem}
\usepackage{amsmath}
\usepackage{wrapfig, lipsum}
\usepackage{comment}

\setcitestyle{numbers,square,comma} 
\usepackage[pagebackref=true,breaklinks=true,colorlinks,citecolor=blue,urlcolor=blue,linkcolor=black,bookmarks=false]{hyperref}
\usepackage[utf8]{inputenc} 
\usepackage[T1]{fontenc}    
\usepackage{hyperref}       
\usepackage{url}            
\usepackage{booktabs}       
\usepackage{amsfonts}       
\usepackage{nicefrac}       
\usepackage{microtype}      
\usepackage{xcolor}         
\usepackage{graphicx}
\usepackage{subcaption}

\title{On the Role of Visual Grounding in VQA}

\author{
  Daniel Reich, Tanja Schultz \\
  Cognitive Systems Lab. \\
  University of Bremen, Germany \\
  \texttt{\{dreich, tanja.schultz\}@uni-bremen.de} \\
}

\begin{document}

\maketitle

\begin{abstract}
Visual Grounding (VG) in VQA refers to a model's proclivity to infer answers based on question-relevant image regions. Conceptually, VG identifies as an axiomatic requirement of the VQA task. In practice, however, DNN-based VQA models are notorious for bypassing VG by way of shortcut (SC) learning without suffering obvious performance losses in standard benchmarks. 
To uncover the impact of SC learning, 
Out-of-Distribution (OOD) tests have been proposed that expose a lack of VG with low accuracy. 
These tests have since been at the center of VG research and served as basis for various investigations into VG's impact on accuracy. However, the role of VG in VQA still remains not fully understood and has not yet been properly formalized. \\
In this work, we seek to clarify VG's role in VQA by formalizing it on a conceptual level. We propose a novel theoretical framework called ``Visually Grounded Reasoning'' (VGR) that uses the concepts of VG and Reasoning to describe VQA inference in ideal OOD testing.
By consolidating fundamental insights into VG's role in VQA, VGR helps to reveal rampant VG-related SC exploitation in OOD testing, which explains why the relationship between VG and OOD accuracy has been difficult to define.
Finally, we propose an approach to create OOD tests that properly emphasize a requirement for VG, and show how to improve performance on them.
\end{abstract}

\section{Introduction}
\label{sec:intro}
Visual Question Answering (VQA) is the task of answering questions about image contents. VQA models process two input modalities to produce an answer: vision (the image) and language (the question). A VQA model's answer inference is called Visually Grounded (VG), if it is based on question-relevant parts of the image. This process is sometimes more intuitively described as being ``Right for the Right Reasons'' \citep{ying2022visfis, right_for_right}. On a conceptual level, the
necessity of VG is clear and its role seems obvious, but in practice, Deep Learning-based (DL) VQA models are notorious for their lack of proper VG \citep{vqacp, agrawal-etal-2016-analyzing_behavior, goyal2017makingv, Han2021GreedyGE, Gupta2022SwapMixDA, reich2022vlr, reich2023fpvg}, while at the same time reaching unprecedented performances in standard benchmarks such as VQA \citep{vqa, goyal2017makingv} and GQA \cite{gqa_dataset}.
This apparent paradox stems from the fact that DL-based models face a significant limitation called shortcut learning \citep{geirhos2020_shortcut_learning}. A shortcut (SC) is characterized as an \textit{unintended solution}\footnote{As an analogy, consider the following two strategies to solve a multiplication task: 1) Looking up results in a fixed table (\textit{unintended solution}), and 2) learning the underlying mathematical concept of multiplication (\textit{intended solution}). The \textit{unintended solution} (i.e., the SC) works well for numbers in the given table but fails for different ones -- which is where the \textit{intended solution} prevails.} to a given problem. 
SCs in DL-based models work well in tests that are created by drawing samples from the same data distribution as the training set (so-called independent and identically distributed tests, ID), but fail to generalize to more challenging conditions, which the \textit{intended solution} does. As a result, superficial performance indicators like accuracy may grossly misrepresent the actual capabilities learned by the model. According to \cite{geirhos2020_shortcut_learning}, conditions that expose SC learning by low accuracy can be implemented by ideal Out-of-Distribution (OOD) tests that have a clear distribution shift away from the training data, and have a well-defined \textit{intended solution} that needs to be applied for success.
In VQA, we intuitively understand that the involvement of relevant image regions is essential to the \textit{intended solution} of the task. Hence, OOD tests present themselves as natural candidates for exposing problematic VG in VQA with low accuracy, but only if the \textit{intended solution} is a requirement for success. 
In VQA, OOD tests created by distribution shifts, such as the VQA-based \cite{vqa} VQA-CP \cite{vqacp} and the data splits in VisFIS \cite{ying2022visfis}, have been introduced for VG research and have taken a central role in related investigations due to VG's presumed importance in generalization scenarios \citep{hint, Wu2019SelfCriticalRF, mutant, shrestha-etal-2020-negative, neural_state_machine, Han2021GreedyGE, reich2022vlr, reich2023fpvg, reich2024truevg}.
Consequently, low(er) accuracy in OOD (vs. ID) testing has been strongly associated with a model's disregard of relevant visual information \cite{vqacp, hint, Wu2019SelfCriticalRF, reich2023fpvg}. 
In line with this association, various works reported success on OOD tests by focusing on VG \cite{vqacp, hint, Wu2019SelfCriticalRF, mutant, reich2022vlr}. However, \cite{shrestha-etal-2020-negative, Han2021GreedyGE} found that VG improvements in \cite{hint, Wu2019SelfCriticalRF} were not actually the source of the reported OOD gains, thereby implying that an association of VG and OOD performance is actually unreliable. 
Consequently, investigations sought to make sense of the nature of VG's role by trying to understand its exact involvement in OOD generalization.
\cite{ying2022visfis} found VG to be a strong predictor of answer correctness in both ID and OOD tests, 
without making an explicit distinction between the two settings. More importantly, \cite{ying2022visfis} showed that the way VG is measured needs to be chosen carefully as it can dramatically affect conclusions regarding VG's impact. 
Subsequently, \cite{reich2023fpvg} introduced an appropriate VG metric and showed that OOD tests benefit more from VG than ID tests, thereby confirming VG's increased importance in generalization. This was similarly reported in \cite{reich2024truevg} which additionally showed that VG impact is heavily influenced by noisy visual features, affecting OOD tests in particular. \cite{reich2024truevg} further found that the quality of VG annotations and the nature of examined questions play a significant role in determining relevant visual information, which is crucial for VG analysis and a potential source for misinterpretations.\\
Collectively, these findings indicate an obscure and complicated relationship between VG and OOD testing that is hard to grasp. 
In this work, we seek to clear up some of the obscurity. 
We propose a novel theoretical framework that establishes VG's role in VQA in a straightforward and comprehensible way with the use of formal logic statements. Our framework, called Visually Grounded Reasoning (VGR), describes the \textit{intended solution} as it relates to three aspects of VQA inference: VG, Reasoning and answer accuracy. 
We further use VGR to reveal that VG is not a strict requirement for success in a number of OOD tests commonly used in VG research, which sheds light on why impact of VG has been difficult to quantify. 
Finally, we propose an approach for creating OOD tests that emphasize the necessity of VG in VQA far better than currently employed tests and investigate how to succeed on them.

\textbf{Contributions.}
Summarized as follows:
\begin{itemize}[noitemsep,nolistsep]
    \item A novel theoretical framework, called Visually Grounded Reasoning (VGR), that formally establishes the involvement of VG in the \textit{intended solution} for the VQA task.
    \item New insights into the role of VG in VQA inference.
    \item An analysis that shows that current OOD tests are an unreliable basis for conclusions regarding VG, because many questions can be answered without it.
    \item An approach to create (and improve performance on) OOD tests that require VG to succeed.
\end{itemize}

\section{Background}
\label{sec:related_work}

\textbf{VG in VQA and its relation to OOD testing.}
With the introduction of the large-scale VQA dataset \citep{vqa} and subsequent influx of monolithic DL-based VQA models, the involvement of relevant image information in VQA inference has become a topic of increasing interest \citep{agrawal-etal-2016-analyzing_behavior, yinyang2016, goyal2017makingv, Johnson2016CLEVRAD}. Lack of VG was prominently shown to be a widespread issue in VQA with the introduction of the OOD split VQA-CP \cite{vqacp}, which promised to expose the problem with low OOD answer accuracy. Similarly constructed tests followed later in \cite{ying2022visfis}.
Various VQA models have since been proposed to facilitate the manifestation of VG through their architectural design \citep{chen2021meta, pvr, reich2022vlr, vqacp, neural_state_machine, neurosymbolic_reasoning_dfol, foundreason}. Special training approaches, such as additional loss functions, that attempt to strengthen VG in existing VQA models have been introduced in \citep{hint, Wu2019SelfCriticalRF, mutant, Han2021GreedyGE, Gupta2022SwapMixDA, ying2022visfis, reich2024truevg}. 
The success of such methods is predominantly determined by their impact on (OOD) accuracy rather than their impact on VG quality, as accuracy is the primary measure for success in VQA. OOD tests that prevent SC exploitation from achieving high accuracy are therefore ideal tools for inspiring the development of models that solve the underlying task instead of running the risk of only solving the dataset. In line with this motivation, various reports have shown that creating OOD tests by changing answer priors, as done in \citep{vqacp, ying2022visfis, kervadec2021roses}, does not manage to break up all spurious correlations found in a dataset and more tests have been proposed to help expose these potential SCs \citep{beyond_question_biases_dancette, explicit_bias_discovery, si-etal-2022-language_is_not_the_only_shortcut}. 
The requirement for the \textit{intended solution} of VG to succeed on these tests, however, has not been explicitly verified in any of the mentioned works. We rectify this in this work for a number of OOD tests using concrete VG analysis. 

\textbf{VG definition.} Insights in \cite{ying2022visfis} imply the importance of clearly outlining the definition of VG in VQA, as it profoundly influences training approaches and the selection of an appropriate metric. In most works in VQA, VG involves identifying a model's \textit{faithful} reliance on image inputs used to produce an answer. However, some works frame VG as a pure localization task where question-relevant image regions are identified by an auxiliary objective for the model \citep{Chen_2022_visuallyImpaired, chen2021meta}. 
This framing of VG is more in line with the definition used in the field of Referring Expressions \citep{kazemzadeh-etal-2014-referitgame}, and falls into the category of \textit{plausible} explanations that lack a concrete connection to the actual outcome of the prediction (i.e., they lack \textit{faithfulness}, see also \cite{jacovi-goldberg-2020-faithfulplausible}). 
This work focuses on the former definition of VG. Additional background on measuring VG in VQA is provided in App. \ref{app:additional_background}.


\section{Visually Grounded Reasoning}
\label{ch8_sec:visually_grounded_reasoning}
We propose a theoretical framework called \textit{Visually Grounded Reasoning} (VGR) that conceptualizes the \textit{intended solution} for VQA.
VGR describes SC-free inference as a dependency between \textit{VG}, \textit{Reasoning} and \textit{answer correctness}. We capture this dependency with formal propositional logic statements. 

\subsection{Reasoning}
\label{subsec:hypothesis_1}
From a conceptual standpoint, the abstract notion of ``Reasoning'' is expected to be central to the \textit{intended solution} of the VQA task. 
This is similarly acknowledged by \cite{corentin_kervadec_thesis} who opts to recognize Reasoning in VQA models by ``what it is not'', and settles on defining it as ``the opposite of exploiting biases and spurious correlation in the training data'' and, therefore, the opposite of exploiting SCs (\cite{corentin_kervadec_thesis}, p. 14, and \cite{kervadec2021transferable, kervadec2021roses}). 
Consequently, \cite{corentin_kervadec_thesis} argues that VQA Reasoning can be quantified as accuracy in OOD tests (\cite{corentin_kervadec_thesis}, p. 15). According to \cite{geirhos2020_shortcut_learning}, the purpose of OOD tests is to reflect a model's successful use of the \textit{intended solution}. Hence, \cite{corentin_kervadec_thesis}'s argument implies the hypothesis that Reasoning is the \textit{intended solution}. 
We formulate this hypothesis in the context of VQA behavior in OOD tests with the following logic statement:

\begin{table}
\centering
\caption{Truth tables for Hypothesis 1 (left) and 2 (right).
Case 2 is invalid in both, i.e., a True answer cannot result from False Reasoning or VG in Hypothesis 1 and 2, respectively.}
\begin{subtable}{.4\columnwidth}
\centering
\small
\caption{Truth table for Hypothesis 1.}
\label{table:answer_reasoning_kervadec}
\begin{tabular}{cccc}
\toprule
Case & Reasoning & Answer & Validity \\
\midrule
1 & \xmark & \xmark & True \\
2 & \xmark & \cmark & False \\ 
[-1.5ex] \hline \\ [-1.5ex]
3 & \cmark & \xmark & True \\
4 & \cmark & \cmark & True \\
\bottomrule
\end{tabular}
\end{subtable}%
\hfill
\begin{subtable}{.5\columnwidth}
\centering
\small
\caption{Truth table for \protect\hyperlink{subsec:hypothesis_2}{Hypothesis 2}.}
\begin{tabular}{ccclc}
\toprule
Case & Visual Grounding & Answer & Validity \\
\midrule
1 & \xmark & \xmark & True \\
2 & \xmark & \cmark & False \\ 
[-1.5ex] \hline \\ [-1.5ex]
3 & \cmark & \xmark & True \\
4 & \cmark & \cmark & True \\
\bottomrule
\end{tabular}
\label{table:answer_vg}
\end{subtable}
\end{table}

\paragraph{Hypothesis 1.}
\begin{equation}
\label{eq:a-re}
\begin{aligned}
    Answer \rightarrow Reasoning \\
\neg Reasoning \rightarrow \neg Answer
\end{aligned}
\end{equation}

In words: A correct answer implicates correct Reasoning. Incorrect Reasoning results in a wrong answer. Table \ref{table:answer_reasoning_kervadec} lists the formal truth table for Hypothesis 1.

\subsection{Visual Grounding}
\label{subsec:hypothesis_2}

VG is an axiomatic requirement of VQA inference by definition of the VQA task, which is to answer questions about image contents. Therefore, like Reasoning, VG has to play a central role in the \textit{intended solution}.
While VG quality has not been explicitly equated to OOD accuracy the way that Reasoning has been in \cite{corentin_kervadec_thesis}, VG quality is deeply associated with OOD accuracy and is investigated based on that premise \citep{vqacp, hint, Wu2019SelfCriticalRF, shrestha-etal-2020-negative, ying2022visfis, reich2023fpvg}. 
Akin to Reasoning, we therefore hypothesize VG's involvement in SC-free inference with a similar logic statement as follows.
\paragraph{Hypothesis 2.}
\begin{equation}
\label{eq:a-vg}
    \begin{aligned}
    Answer \rightarrow VG \\
    \neg VG \rightarrow \neg Answer
    \end{aligned}
\end{equation}

In words: A correct answer implicates correct VG. Incorrect VG results in a wrong answer.
The truth table for \hyperlink{subsec:hypothesis_2}{Hypothesis 2} is shown in Table \ref{table:answer_vg}.

\subsection{The VGR Proposition}
Conceptually, both Reasoning and VG are axiomatic requirements of the \textit{intended solution} for VQA. We therefore combine the two hypotheses to describe SC-free VQA inference as follows.

\paragraph{VGR Proposition.}
\begin{equation}
\label{eq:a-re-vg}
\begin{aligned}    
    Answer \rightarrow Reasoning \wedge VG \\
     \neg (Reasoning \wedge VG) \rightarrow \neg Answer
     \end{aligned}
\end{equation}

In words: A correct answer implicates both, correct Reasoning and VG. Without both, correct Reasoning and VG, the answer cannot be correct. Table \ref{table:VGR_proposition} lists the formal truth table for the VGR Proposition, which we discuss in detail in Sec. \ref{model_behavior_VGR_FPVG}.

\begin{table}[t]
\centering
\caption{All 8 cases of SC-free inference described by Reasoning, VG and answer correctness. Each case's validity is listed for Hypothesis 1 and 2 and the VGR Proposition (strikethrough lines represent invalid cases under VGR). Final column lists corresponding categorization with FPVG (see Sec. \ref{model_behavior_VGR_FPVG}).}
\resizebox{1\textwidth}{!}{%
\begin{tabular}{cccccccc}
\toprule
& Reasoning & Visual Grounding & Answer & \hyperlink{subsec:hypothesis_1}{Hypothesis 1} & \hyperlink{subsec:hypothesis_2}{Hypothesis 2} & VGR Proposition & \\
Case & (RE) & (VG) & (A) & $A\rightarrow RE$ & $A \rightarrow VG$ & $A\rightarrow RE \wedge VG$ & FPVG \\
\midrule
1.1 & \xmark & \xmark & \xmark & True & True & True & BGW \\
1.2 & \xmark & \cmark & \xmark & True & True & True & GGW \\
2.1 & \xmark & \xmark & \cmark & False & False & False & \textbf{BGC} \\
[-1.5ex] \hline \\ [-1.5ex]
2.2 & \xmark & \cmark &  \cmark & False & True & False & GGC \\
[-1.5ex] \hline \\ [-1.5ex]
3.1 & \cmark & \xmark &  \xmark & True & True & True & BGW \\
3.2 & \cmark & \cmark &  \xmark & True & True & True & GGW \\
4.1 & \cmark & \xmark &  \cmark & True & False & False & \textbf{BGC} \\
[-1.5ex] \hline \\ [-1.5ex]
4.2 & \cmark & \cmark &  \cmark & True & True & True & GGC \\
\bottomrule
\end{tabular}
 }
\label{table:VGR_proposition}
\end{table}

\subsection{Hypothesis 1 \& 2 are insufficient as individual descriptions of SC-free inference}
\label{ch8_subsec:hypo1_flawed}

\hyperlink{subsec:hypothesis_1}{Hypothesis 1} was presented as a self-sufficient description of the \textit{intended solution} in \cite{corentin_kervadec_thesis}. Since VG is an axiomatic requirement that has to be involved in the \textit{intended solution}, self-sufficiency would only be confirmed if VG was conflated with Reasoning in \hyperlink{subsec:hypothesis_1}{Hypothesis 1}.
We show why this is not possible. 
Table \ref{table:VGR_proposition} lists all eight permutations of the three involved aspects of Reasoning, VG and answer correctness. Assuming that Reasoning and VG are conflated, we find that under \hyperlink{subsec:hypothesis_1}{Hypothesis 1} 
Case 4.1 (i.e., a correct answer given based on correct Reasoning but incorrect VG) would represent a valid case in SC-free inference. However, Case 4.1 \textit{refutes the axiomatic involvement of VG in VQA}. 
Concretely, under \hyperlink{subsec:hypothesis_2}{Hypothesis 2} Case 4.1 is in fact considered a SC, as signified by its invalidity. Hence, we surmise that \textit{VG cannot be conflated with Reasoning} 
and must be explicitly considered alongside it as a separate component on equal footing. \\
A similar argument can be made to show the insufficiency of \hyperlink{subsec:hypothesis_2}{Hypothesis 2} (Case 2.2 refutes the axiomatically required involvement of Reasoning).

\subsection{Summary of Insights}
VGR consolidates the following insights regarding VG's role in SC-free inference: 
\begin{enumerate}[noitemsep,nolistsep]
    \item Correct VG / Reasoning \textit{on its own} does not describe the \textit{intended solution}.
    \item VG and Reasoning \textit{cannot be conflated}, but are complementary influences in VQA inference.
    \item Both correct VG and Reasoning \textit{have to coincide} for a correct answer.
\end{enumerate}
Notably, this has the following implications for VG-related OOD analysis:
\begin{itemize}[noitemsep,nolistsep]
    \item VG improvements do not have to translate to OOD accuracy improvements
    \item OOD accuracy improvements do not have to mean VG has improved
\end{itemize}

\section{VGR in Practice}
\label{model_behavior_VGR_FPVG}

VGR provides an abstract formulation of the \textit{intended solution} in VQA. 
To bridge the gap from theory (VGR) to practice (observed inference in OOD testing), we use the VG metric ``FPVG'' \citep{reich2023fpvg}. 
In FPVG, every evaluated question is assigned one of four categories based on joint-measurements of answer correctness (correct or wrong) and VG (good or bad). These four categories are summarized by two overarching VG-centric categories which disregard answer correctness, $FPVG_{+}$ (good VG) and $FPVG_{-}$ (bad VG). All FPVG categories and their meaning are listed in Table \ref{table:fpvg_categories} for reference.

\begin{wraptable}{r}{0.45\textwidth}
\begingroup
\setlength{\tabcolsep}{4pt}
\small
\vspace{-12pt}
    \caption{FPVG categories.}
    \vspace{-8pt}
    \begin{tabular}{ll}
    \toprule
         GGC & \textbf{G}ood \textbf{G}rounding, \textbf{C}orrect Answer \\
         GGW & \textbf{G}ood \textbf{G}rounding, \textbf{W}rong Answer \\
         BGC & \textbf{B}ad \textbf{G}rounding, \textbf{C}orrect Answer \\
         BGW & \textbf{B}ad \textbf{G}rounding, \textbf{W}rong Answer \\
         \midrule
          $FPVG_{+}$ & Good Grounding (GGC + GGW) \\
          $FPVG_{-}$ & Bad Grounding (BGC + BGW) \\
         \midrule
          Accuracy & GGC + BGC \\
    \bottomrule
    \end{tabular}
    \label{table:fpvg_categories}
    \vspace{-9pt}
\endgroup
\end{wraptable}



We map VGR's \textit{theoretical} eight cases of inference in Table \ref{table:VGR_proposition} onto \textit{measurable} FPVG categories listed in Table \ref{table:fpvg_categories} (top four entries). The mappings are listed in Table \ref{table:VGR_proposition} under column ``FPVG'' (e.g., 
``GGC'' represents VGR's Case 4.2). 
Based on this mapping, we can identify certain patterns that allow us to verify whether the VGR-defined \textit{intended solution}
is used in OOD testing or not. 
We formulate these patterns as the following corollaries.

\subsection{VGR corollaries of SC-free inference}
\label{subsec:4corollaries}

\paragraph{Corollary 1: BGC is zero.}
\label{subsec:corollary1}
We find that BGC (i.e., questions that evaluate as correctly answered despite bad VG) has no valid match in VGR (see Table \ref{table:VGR_proposition}, final column). While Cases 2.1 and 4.1 in Table \ref{table:VGR_proposition} match the conditions for BGC, they are invalid under the VGR Proposition.
We can infer from this that \textit{BGC captures cases of SC exploitation}. Consequently, the share of questions that are identified as BGC should ideally be zero
in an OOD test, regardless of the tested VQA model:
\begin{equation}
\begin{split}
\label{eq:bgc_zero}
BGC & = 0 \\
\end{split}
\end{equation}
\paragraph{Corollary 2: GGC equals Accuracy.}
\label{subsec:corollary2}
Since $BGC = 0$ in OOD tests, we can reformulate FPVG's formula for accuracy (Table \ref{table:fpvg_categories}, bottom line) as follows:
\begin{equation}
\begin{split}
\label{eq:acc_ggc}
Acc & = GGC + BGC \\
& = GGC \\
\end{split}
\end{equation}



\paragraph{Corollary 3: Accuracy cannot surpass $\mathbf{FPVG_{+}}$.}
\label{subsec:corollary3}
By applying \hyperlink{subsec:corollary2}{Corollary 2} to FPVG's formula of VG (i.e., $FPVG_{+}$ in Table \ref{table:fpvg_categories}), we find that models cannot achieve higher accuracy than $FPVG_{+}$. 
\begin{equation}
\begin{split}
\label{eq:fpvg_acc}
FPVG_{+} & = GGC + GGW \\
& = Acc + GGW \\
& \geq Acc \\
\end{split}
\end{equation}


\subsection{Limitation: Theory vs. practice}
\label{subsec:theoryVSpractice}
The VGR corollaries describe expected results in OOD testing assuming ideal, fully controlled testing conditions. 
Such ideal conditions are unlikely to be fully enforced in practical testing and therefore a small degree of transgressions of the corollaries 
are to be expected and may be unavoidable. E.g., we do not expect BGC (Corollary 1) to be \textit{exactly zero} in practice. 
Formally, the VGR Proposition does not account for the impact of deviations from ideal testing conditions that are encountered in practice, as these are not straightforward to predict and quantify. Therefore, when using VGR for analysis of a given test set in practice, we recommend considering how closely the corollaries are approximated, rather than verifying strict and exact observance, when determining their violation.

\section{Do current OOD tests measure SC-free VQA performance?}
\label{section:existing_ood_tests}
OOD tests used for evaluating the \textit{intended solution} should not be solvable by SC exploitation (cf. \citep{geirhos2020_shortcut_learning}). 
We examine results of four current OOD tests and compare them with VGR-defined behavior to verify whether or not they are proper OOD tests.

\begin{wraptable}{r}{0.47\textwidth}
\begingroup
\small
\caption{Dataset sizes.}
\label{table:gqa_ood_dataset}
\setlength{\tabcolsep}{4pt}
\begin{tabular}{lrrrr}
\toprule
Dataset & Train & Dev & ID & OOD \\
\midrule
GQA-CP-large \citep{ying2022visfis} & 645k & 107k & 139k & 137k \\
GQA-OOD \citep{kervadec2021roses} & 923k & 20k & 29k & 15k  \\
VQA-CPv2 \citep{vqacp} & 418k & 20k & n/a & 15k \\
VQA-HAT-CP \citep{ying2022visfis} & 36k & 6k & 7.2k & 7.0k \\
\bottomrule
\end{tabular}
\endgroup
\end{wraptable}
\subsection{Experiment Preliminaries}
\label{section:experiment_preliminaries}
\textbf{Datasets.}
We evaluate four OOD dataset splits that are based on GQA \citep{gqa_dataset} and the VQA dataset \cite{vqa, goyal2017makingv}. GQA provides VG annotations used for measuring VG with FPVG for the majority of questions. For measuring VG on the VQA dataset we use VG annotations provided by VQA-HAT \cite{Das2016HumanAI}, which annotates a small percentage of questions in the VQA dataset. GQA-CP-large and VQA-HAT-CP \cite{ying2022visfis} are splits that were created by re-distribution of questions in GQA and VQA-HAT, respectively. Similarly, VQA-CPv2 is a re-distribution of VQAv2 \citep{goyal2017makingv}. All three ``CP'' splits were created following the ``Changing Priors'' approach described in \cite{vqacp}, which disaligns answer prior distributions for every question type in train and test set. 
GQA-OOD \citep{kervadec2021roses}, which was introduced for measuring Reasoning abilities (which implicitly includes VG), keeps GQA's original train set, but redistributes val questions into ID/OOD based on frequency of answers per question type. \\
Sample counts are listed in Table \ref{table:gqa_ood_dataset}. Testing only involves questions that have meaningful VG annotations, as both VG measurements and answer correctness are needed for each question to perform the analysis.


\textbf{VQA Models.}
Initial experiments involve two VQA models: \textbf{UpDn} \citep{bottomup_paper}, a classic, single-hop attention-based model, and \textbf{LXMERT} \citep{lxmert}, a Transformer-based \citep{attentionisallyouneed}, BERT-like model \citep{devlin2018bert} trained under a pre-train/fine-tune paradigm. \\
Each model is trained five times with a different random seed for each data split. LXMERT's pre-training is performed for each dataset to uphold the intended sample distributions. LXMERT is not evaluated with VQA-HAT-CP due to its small size. 
App. \ref{appsec:model_training_details} has additional training details.

\textbf{Visual Features.}
All VQA-based models use 2048-dim object-based visual features from \cite{bottomup_paper}. All GQA-based models use object-based \textit{symbolic} visual features as image representation, shared by \cite{reich2024truevg}. 
These are constructed by concatenation of GloVe word embeddings \citep{glove} describing an object's name, attributes and location in the image (illustration in Fig. \ref{figure:vgr_augmentation}, left). Additional details can be found in App. \ref{app:visual_features}.

\begin{table*}[t]
\centering
\caption{Accuracy and FPVG results for three OOD tests, evaluated with UpDn and LXMERT. All results are averaged over five differently seeded runs (result ranges are listed in App. \ref{app:additional_results_ood}). 
} 
\resizebox{\textwidth}{!}{%
\begin{tabular}{lcccccccccccccccc}
\toprule
 \multicolumn{2}{c}{OOD Training} & \multicolumn{7}{c}{ID} && \multicolumn{7}{c}{OOD} \\
 \cmidrule{3-9}
 \cmidrule{11-17}
 Dataset & Model & Acc & $FPVG_{+}$ && GGC & GGW & BGC & BGW && Acc & $FPVG_{+}$ && GGC & GGW & BGC & BGW \\
\midrule
 GQA-CP-large & UpDn & 64.53 & 24.30 && 18.65 & 5.65 & 45.89 & 29.82 && 44.60 & 23.46 && 14.94 & 8.52 & 29.66 & 46.87 \\
 & LXM & 70.05 & 24.13 && 20.16 & 3.97 & 49.89 & 25.98 && 53.51 & 23.69 && 17.29 & 6.40 & 36.22 & 40.09 \\
\midrule
GQA-OOD & UpDn & 63.18 & 27.52 && 20.95 & 6.58 & 42.23 & 30.25 && 43.72 & 26.78 && 15.82 & 10.97 & 27.90 & 45.32 \\
& LXM & 65.54 & 26.59 && 21.31 & 5.28 & 44.23 & 29.18 && 47.09 & 25.08 && 16.06 & 9.02 & 31.03 & 43.89 \\
\midrule
VQA-CPv2 & UpDn &  &  && n/a &  &  &  && 41.53 & 23.54 && 14.16 & 9.38 & 36.15 & 40.31 \\
 & LXM &  &  && n/a &  &  &  && 42.24 & 17.44 && 11.30 & 6.15 & 39.81 & 42.74 \\
\midrule
VQA-HAT-CP & UpDn & 54.49 & 23.14 && 12.89 & 10.25 & 46.52 & 30.34 && 40.80 & 26.57 && 11.83 & 14.74 & 33.94 & 39.49 \\
\bottomrule
\end{tabular}
}  
\label{table:ood_tests}
\end{table*}

\subsection{Result Discussion}
\label{ch8_subsec:current_OOD_tests_discussion}
The examined OOD tests are intended to uncover SC exploitation and reflect a model's VG and Reasoning capabilities. Therefore, OOD results should align closely with VGR. Remarkably, however, results in Table \ref{table:ood_tests} show that \textit{all examined OOD tests violate the VGR corollaries}:

\textbf{\hyperlink{subsec:corollary1}{Corollary 1}}: BGC represents a \textit{substantial share} of questions, when it should approximate \textit{zero}. \\ 
\textbf{\hyperlink{subsec:corollary2}{Corollary 2}:} GGC is considerably \textit{lower} than Accuracy, when it should be in \textit{similar range}. \\
\textbf{\hyperlink{subsec:corollary3}{Corollary 3}:} $FPVG_{+}$ is far \textit{lower} than Accuracy, when it should be \textit{similar or higher}.

Hence, according to VGR, none of these three OOD tests quantify SC-free behavior. 
VG-related SC exploitation, in particular, is rampant in these tests which emphasizes that OOD accuracy does not provide reliable evidence that a model has learned to rely on VG to succeed.
Furthermore, the fact that correct answers can be achieved with \textit{and without} VG, even in OOD tests, offers a clear explanation \textit{why} accuracy and VG have an unpredictable relationship (cf. \cite{shrestha-etal-2020-negative, ying2022visfis, reich2023fpvg}). 
In this context, we note that OOD tests do indicate a move in the right direction by offering less SC opportunities than ID tests, which is evident when comparing respective BGC numbers in Table \ref{table:ood_tests}. A similar observation was made in \citep{reich2023fpvg}, where OOD accuracy was found to be more sensitive to VG than ID accuracy.

\section{Creating VGR-conforming OOD tests using augmentation}
\label{ch8_sec:aug-ood-creation-steps}
\begin{figure*}[t]
\includegraphics[width=0.99\columnwidth]{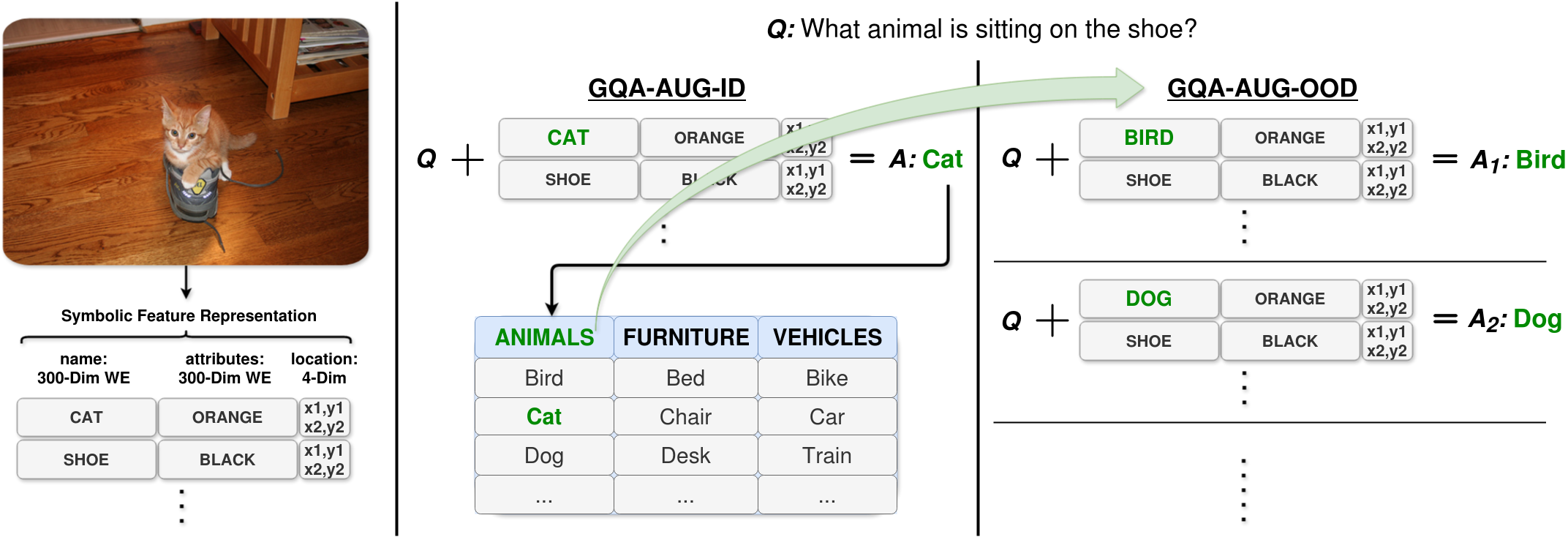}
\caption{Examples of symbolic visual features and samples in GQA-AUG. Left: Symbolic feature representation. AUG-ID (middle) shows an original test sample (question and visual features) with the ground-truth answer (``cat''). AUG-OOD (right) contains new samples that differ in both answer (``dog'', ``bird'', ...) and  feature content (modified to support the answer). The question is not changed.} 
\label{figure:vgr_augmentation}
\end{figure*}
The investigated OOD tests in Sec. \ref{section:existing_ood_tests} were created by controlling Q/A prior distributions. This approach does not explicitly account for VG-related SC exploitation.
We outline an approach for creating \textbf{GQA-AUG} (\textit{AUG}mentation), an OOD test split that only includes questions that are unlikely to be answered correctly without reliance on correct VG. 
We create GQA-AUG using the following steps (illustration in Fig. \ref{figure:vgr_augmentation}):
\begin{enumerate}[noitemsep,nolistsep,itemsep=0pt, wide=0pt, leftmargin=*, after=\strut]
    \item Identify \textit{query}-type questions in GQA's balanced val set with answers consisting of object names (e.g., dog, car, etc.) tied to an object in the image. We call this set of questions \textbf{AUG-ID}.
    \item Identify the object's category (e.g., animal, vehicle, etc.). 
    \item Generate \textit{new Q/A samples} by replacing the answer with up to ten uniquely sampled names from the same object category (e.g., cat $\rightarrow$ dog, bird, etc.). The question remains the same.
    \item Modify the originally detected \textit{image representation} for each new Q/A sample to support the new answer. Concretely, we modify relevant feature vectors in the image representation such that their feature content supports the new answer (Fig. \ref{figure:vgr_augmentation}, right). 
    We call this set of questions \textbf{AUG-OOD}.
\end{enumerate}
\begin{wraptable}{r}{0.49\textwidth}
\begingroup
\setlength{\tabcolsep}{4pt}
\small
\vspace{-12pt}
\centering
\caption{Sample counts for GQA-AUG.}
\begin{tabular}{lrrrr}
\toprule
Dataset & Train & Dev & AUG-ID & AUG-OOD \\
\midrule
GQA-AUG & 829k & 20k & 16k & 156k \\
\bottomrule
\end{tabular}
\label{table:gqa_aug_dataset}
\vspace{-12pt}
\endgroup
\end{wraptable}

This approach of creating new \textit{query}-type questions alongside visual feature augmentation is intended to minimize the possibility of correct answers being returned without correct VG. Additionally, step 3 introduces a shift of answer priors.
Dataset numbers for GQA-AUG are listed in Table \ref{table:gqa_aug_dataset}, additional statistics are given in App. \ref{app:additional_details_GQAAUG}.

\section{Experiments on GQA-AUG}
\label{sec:experiments_on_gqa_aug}
\subsection{Used VQA Models}
We evaluate five VQA models of different architectural designs:
\textbf{MAC} \citep{mac} is a multi-hop attention-based model developed for GQA-type Visual Reasoning. \textbf{MMN} \citep{chen2021meta} is a Transformer-based model which uses question programs generated by an independently trained question parser instead of the otherwise common word embeddings used for raw question input. \textbf{VLR} \citep{reich2022vlr} uses rule-based inference that approximates the \textit{intended solution} for Information Retrieval-type (IR) questions (such as the \textit{query}-questions in GQA). \textbf{UpDn} and \textbf{LXMERT} have been introduced in Sec. \ref{section:existing_ood_tests}.
Training details and additional description for all models can be found in App. \ref{appsec:model_training_details}.

\subsection{Does AUG-OOD measure SC-free VQA performance?}
\label{ch8_subsec:augood_verification_vgr}
Table \ref{table:INF_AUG_allmodels} lists results for evaluations on GQA-AUG. 
We validate the VGR corollaries as follows:

\textbf{Corollary 1 (Low BGC):}
All models in Table \ref{table:INF_AUG_allmodels} post low numbers in BGC for AUG-OOD. In particular, BGC is substantially lower than in other OOD tests. \\ 
\textbf{Corollary 2 (Accuracy is equal to GGC):}
While accuracy is still not equal to GGC due to some residual BGC, \hyperref[subsec:corollary2]{Corollary 2} is far better approximated than in other examined OOD tests. \\
\textbf{Corollary 3 (Accuracy is not higher than $\mathbf{FPVG_{+}}$):}
\hyperref[subsec:corollary3]{Corollary 3} is met by all models except VLR which exceeds $FPVG_+$ by a small margin. We consider this within acceptable range (see discussion in Sec. \ref{subsec:theoryVSpractice}). A non-ideal BGC value (exceeding zero) observed for all models is contributing to this value. 
It is also worth pointing out that AUG-ID results show violation of \hyperref[subsec:corollary3]{Corollary 3} in all models, reaffirming that the original set of questions does not measure SC-free performance.

\textbf{Summary}
AUG-OOD approximates the VGR-derived corollaries to a substantially higher degree than other examined OOD tests. In conclusion, we find AUG-OOD to be a significantly better candidate for measuring SC-free performance.

\begin{table*}[t]
\centering
\caption{GQA-AUG: Results for five models show that AUG-OOD conforms to VGR. Four models trained w/ and w/o Infusion. 
All numbers are averages of five differently seeded runs (result ranges are listed in App. \ref{app:additional_results_aug}).
VLR uses rule-based inference (=not trained). Infusion training significantly improves accuracy and $FPVG_+$ on AUG-OOD, while VGR is upheld. Discussion in Sec. \ref{sec:experiments_on_gqa_aug}.} 
\resizebox{\textwidth}{!}{%
\begin{tabular}{lcccccccc}
\toprule
 & & \multicolumn{2}{c}{Accuracy ($FPVG_+$)} && \multicolumn{4}{c}{FPVG for AUG-OOD} \\
 \cmidrule{3-4}
 \cmidrule{6-9}
 Model & Infusion \cite{reich2024truevg} & AUG-ID & AUG-OOD && GGC & GGW & BGC & BGW  \\
\midrule
UpDn \cite{bottomup_paper} & no & 40.09 (33.99) & 16.27 (27.27) && 12.57 & 14.70 & 3.70 & 69.03 \\
LXMERT \cite{lxmert} & no & 41.95 (30.78) & 13.79 (18.93) && 8.96 & 9.97 & 4.83 & 76.24 \\
MMN \cite{chen2021meta} & no & 43.09 (40.10) & 21.07 (27.01) && 16.94 & 10.07 & 4.13 & 68.86 \\
MAC \cite{mac} & no & 40.44 (31.64) & 15.79 (21.58) && 9.83 & 11.75 & 5.96 & 72.47 \\
\midrule
VLR \cite{reich2022vlr} & n/a & 39.08 (47.21) & 81.23 (78.76) && 77.26 & 1.50 & 3.97 & 17.27 \\
\midrule
UpDn \cite{bottomup_paper} & yes & 41.67 (42.86) & 64.36 (66.33) && 61.42 & 4.90 & 2.94 & 30.74 \\
LXMERT \cite{lxmert} & yes & 42.17 (41.35) & 58.66 (57.03) && 54.44 & 2.59 & 4.22 & 38.75 \\
MMN \cite{chen2021meta} & yes & 43.77 (43.70) & 59.86 (57.92) && 55.60 & 2.32 & 4.26 & 37.82 \\
MAC \cite{mac} & yes & 40.83 (42.83) & 63.91 (64.48) && 61.06 & 3.42 & 2.85 & 32.67 \\
\bottomrule
\end{tabular}
}  
\label{table:INF_AUG_allmodels}
\end{table*}

\subsection{Result Discussion}
Results for all five models in Table \ref{table:INF_AUG_allmodels} (top 5 lines) support our conclusion about AUG-OOD's conformity to VGR. 
We further find relatively high GGW numbers compared to GGC in most models. In the context of VGR, we interpret this as an indication of underdeveloped Reasoning capabilities to solve the task by involvement of VG, i.e., correct VG exists but does not always translate to correct answers due to a lack of Reasoning. 
VLR is the only model in the table that was designed specifically to implement the \textit{intended solution} under an IR-based paradigm, which is reflected by its exceptional success on AUG-OOD. 
We interpret the contrasting results between VLR and the other four models as an indicator that the four models have forgone the adoption of the \textit{intended solution} in favor of learning to exploit SCs -- which is what we expect an OOD test to reflect by accuracy. In other words, AUG-OOD is working as intended. 

\subsection{Improving performance on GQA-AUG}

\subsubsection{Learning visually grounded reasoning}
\label{ch8_subsec:learning_ir_reasoning}

GQA can be categorized as an IR-type VQA dataset, as the vast majority of its questions are generated by filling in question-templates with explicit (retrievable) information taken from annotated scene graphs of involved images. Therefore, we see no obvious dataset-related reasons preventing a model to learn IR-type Reasoning to succeed on AUG-OOD. 
Consequently, we look for the problem's source elsewhere. 
\cite{kervadec2021transferable} reported insights that noisy visual inputs interfere with a model's adoption of reasoning patterns. Similarly, a VG-focused analysis in \citep{reich2024truevg} showed that consistently providing accurate, relevant visual targets in training significantly improves VG. These findings suggest that the adoption of IR-type Reasoning, which relies on correct VG, might be impeded by noisy and therefore inconsistently presented visual targets in the input.
We therefore evaluate the training method called ``Information Infusion'' \citep{reich2024truevg} which minimally modifies (``infuses'') the visual input such that it consistently provides accurate question-relevant image content. Keeping with the terminology in \citep{reich2024truevg}, we call regularly trained models ``DET'' and Infusion-trained models ``INF''. 

\subsubsection{Result Discussion}
\label{ch8_subsec:learning_ir_reasoning_result_discussion}

\begin{figure*}[t]
\begin{subfigure}[b]{0.48\textwidth}
    \includegraphics[width=\textwidth]{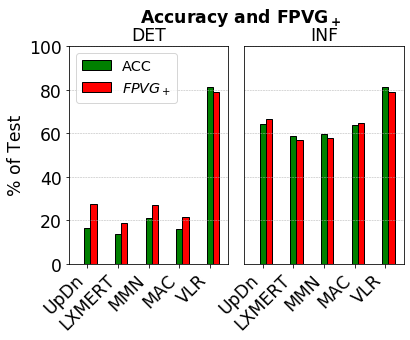}
\end{subfigure}
\begin{subfigure}[b]{0.48\textwidth}
    \includegraphics[width=\textwidth]{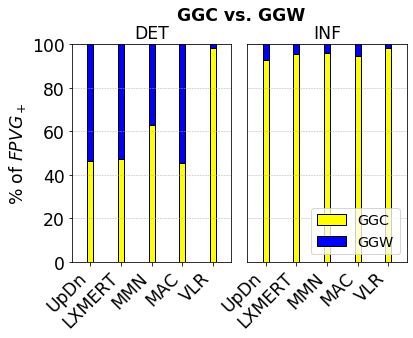}
\end{subfigure}
\hfill
 \vspace{-2mm}
      \caption{Illustration of AUG-OOD results for DET and INF-trained models from Table \ref{table:INF_AUG_allmodels} (averages over five differently seeded runs). Discussion in Sec. \ref{ch8_subsec:learning_ir_reasoning_result_discussion}.}
     \label{vgr_fig:augood_acc_ggc}
\end{figure*}

Numerical results for INF models are listed in Table \ref{table:INF_AUG_allmodels}, bottom four lines. We highlight some AUG-OOD results in Figure \ref{vgr_fig:augood_acc_ggc} and make the following observations.

\textbf{AUG-OOD accuracy is greatly improved.}
Fig. \ref{vgr_fig:augood_acc_ggc}, left, shows that all four re-trained models see substantial gains in AUG-OOD accuracy (green) and $FPVG_+$ (red), while AUG-ID accuracy (not pictured, listed in Table \ref{table:INF_AUG_allmodels}) remains mostly stable. 
In AUG-OOD, the answer and the content of question-relevant visual objects are aligned by design (see Fig. \ref{figure:vgr_augmentation}), therefore improvements in Reasoning and VG are reflected very clearly. 
It is worth noting that INF-training does not involve any changes to the training set's Q/A prior distribution. 
That is to say, improvements in AUG-OOD are not the result of any answer distribution shifts in the training data. \\  
\textbf{GGC has increased while GGW has shrunk.}
Fig. \ref{vgr_fig:augood_acc_ggc}, right, shows that in INF-trained models, GGC is considerably higher than GGW and is approaching $FPVG_+$. Additionally, $FPVG_+$ (and therefore GGC) aligns better with accuracy in INF models than in DET models (notice the differences between red and green bars per model in DET vs. INF in Figure \ref{vgr_fig:augood_acc_ggc}, left). 
We interpret the more effective involvement of VG (as reflected by increased dominance of GGC over GGW in Fig. \ref{vgr_fig:augood_acc_ggc}, right) as evidence of improved Reasoning: The models have learned to leverage relevant visual content (i.e., VG) to produce correct answers.\\
\textbf{VGR corollaries are better approximated.}
Overall, results show that the VGR corollaries are even better approximated after INF training, providing additional validation of AUG-OOD's suitability as OOD test under VGR.

\section{Conclusion}
\label{vgr_sec:conclusion}
We have introduced VGR, a theoretical framework that formally describes the role of VG in VQA inference in the context of SC-learning. 
VGR consolidates a number of significant insights into VG's role in VQA. 
Discussions of empirical results throughout this work illustrate how VGR can help us better understand model behavior in practice, and how this understanding can be leveraged to identify and target shortcomings of the model.
Using VGR, we have shown that current OOD tests still allow many SCs to succeed. Based on this finding, we have proposed a method to create tests that require a model's reliance on VG and have shown how to train models to succeed on them.

\subsection{Closing remarks}
The extent of SC learning in VQA models can be exposed by the employment of properly designed OOD tests that evaluate for a model's adoption of the \textit{intended solution} of the underlying task. Ideal OOD tests, that require the \textit{intended solution} for success, can act as prime indicators of a model's projected generalization capabilities. 
We showed that VG -- even though it is an integral part of the \textit{intended solution} of the VQA task as defined by VGR -- is neither a validated nor enforced requirement for success in current OOD tests used in VG-related research. 
While it is not feasible to exhaustively define \textit{all} decision rules that make up the \textit{intended solution} of a complex task like VQA, we believe that OOD tests should make an effort to involve those decision rules (here: VG) that \textit{can} be defined in specific scenarios. 

\bibliography{nips2024}

\clearpage
\appendix

\section{Dataset License} We conduct experiments on three datasets: GQA \cite{gqa_dataset}, VQA \citep{vqa, goyal2017makingv} and VQA-HAT \cite{Das2016HumanAI}, all under the CC BY 4.0 license.

\section{Additional Background}
\label{app:additional_background}
\subsection{VG metrics} \cite{shrestha-etal-2020-negative} demonstrates the need for VG metrics to prevent misinterpretations of VQA results. However, correctly measuring VG in VQA is considerably more complex than measuring accuracy. 
While some works report VG measurements with a variety of makeshift metrics \citep{gqa_dataset, hint, shrestha-etal-2020-negative, Han2021GreedyGE, reich2022vlr, foundreason}, they share a lack of fundamental research into the metrics' crucial property of \textit{faithfulness} of measurements (i.e., how accurately they capture a model's internal feature importance (FI) for its answer, see also \citep{jacovi-goldberg-2020-faithfulplausible} for this terminology). Experiments in \citep{ying2022visfis} showed that the property of \textit{faithfulness} is crucial when drawing conclusions involving VG. Hence, \citep{ying2022visfis} investigated the \textit{faithfulness} of methods used to determine FI in VQA models, most notably attention \citep{attention}, GradCAM \citep{gradcam} and feature modulation techniques \citep{deyoung-etal-2020-eraser, LOOKOO}, showing the latter type to deliver the most \textit{faithful} measurements. Subsequently, FPVG, a ``Faithful and Plausible'' VG metric based on feature modulation tailored to the VQA task was proposed in \cite{reich2023fpvg}. We use FPVG throughout this work to investigate VQA inference. 

\subsection{VG annotations}
\label{app:background:vg_annotations}
VG annotations of question-relevant image regions are required to determine how \textit{plausible} a model's VG is during inference (i.e., if a model relies on image regions that were determined to be plausibly relevant to the question). 
\textbf{VQA-HAT} \cite{Das2016HumanAI} provides such annotations as human-generated image heat-maps that mark question-relevant image regions for a small percentage of questions in VQAv1 \citep{vqa}. These heat-maps were collected by tracking a user's mouse movements across the image while answering a given question. VQA-HAT's VG annotations are used in our experiments for testing \textbf{VQA-CPv2} and \textbf{VQA-HAT-CP}.\\
\textbf{GQA} \cite{gqa_dataset} (and similar types of datasets, such as \cite{clevr_xai}) provides detailed VG annotations that point out question-relevant objects in the underlying scene graph annotations. Annotations are provided for the majority of questions in the dataset. All GQA-based datasets make use of these VG annotations in our experiments. \\
While not used in this work, there are also computational approaches that attempt to determine question-relevant image regions in the VQA dataset by mapping image annotations and question words \cite{mutant}, or by leveraging textual explanations for the answer \citep{Wu2019SelfCriticalRF}.

\section{Experiment Details}

\subsection{Additional details on GQA-AUG}
\label{app:additional_details_GQAAUG}
Dataset numbers for the GQA-AUG test splits are listed in Table \ref{table:gqa_aug_dataset}. 
AUG-ID consists of 16k unmodified \textit{query}-type questions taken from the GQA balanced val split. Based on AUG-ID, we synthesize 156k new samples using the augmentation process described in Sec. \ref{ch8_sec:aug-ood-creation-steps} which generates up to ten samples per question in AUG-ID, based on the number of unique object names in the involved object category.
On average, AUG-OOD modifies 4.2 query-related objects per question. The average of all question-relevant objects per question in AUG-OOD is 6.6. In 35.7\% of questions in AUG-OOD the set of question-relevant objects overlaps fully with the set of modified objects.

\subsection{Visual Features}
\label{app:visual_features}
VQA and GQA use different sources for their visual features, which we detail below.

\paragraph{VQA} All \textbf{VQA}-based evaluations use regular 2048-dim sub-symbolic visual features generated by a Faster R-CNN \citep{fasterrcnn} object detector shared by and described in \cite{bottomup_paper}. 36 objects (i.e., 36 feature vectors) are extracted per image.

\paragraph{GQA} All \textbf{GQA}-based evaluations use 600-dim symbolic visual features shared by and explained in detail in \cite{reich2024truevg}. The underlying scene graph detector is described in detail in \citet{reich2022vlr}. Up to 100 objects (i.e., 100 feature vectors) are extracted per image. \\
GQA-based experiments involve symbolic visual features which are created and modified as follows (see \cite{reich2024truevg} for a more detailed description): 
\begin{itemize}[noitemsep,nolistsep]
    \item \textbf{DET:} \textit{DET}ection features are assembled based on the recognition output of a scene graph generator. The detected object name and object attributes are each converted into a 300D GloVe word embedding and concatenated with the detected object coordinates (4-dim) to form a 600D object-based feature vector (see also illustration in Fig. \ref{figure:vgr_augmentation}, left). 
    \item \textbf{INF:} Infusion training modifies DET features so that all question-relevant objects mentioned in GQA's VG annotations are accurately represented. E.g., if a relevant object name was misrecognized in DET features, Infusion modifies the object's name information in the vector (first 300-dims in 604-dim vector) by replacing it with the correct name.
\end{itemize}

\subsection{Evaluation details} 

\subsubsection{Sample selection}
\label{app:sample_selection}
Test sets contain a reduced number of samples compared to their full original releases. This is because only a subset of all samples qualify for VG measurements by FPVG. A question must be accompanied by visual features with both relevant and irrelevant objects (based on the definition in FPVG). 
All reported results as well as the test set numbers in the dataset tables account for this reduction.

Similarly, to ensure a clean comparison between DET and INF-trained models in Sec. \ref{sec:experiments_on_gqa_aug}, GQA-AUG's training set is reduced to a subset of samples in GQA's ``balanced'' training set that has VG annotations which allow Infusion training.

\subsubsection{Accuracy and FPVG measurements}
\label{app:accuracy_fpvg_measurements}
VQA and GQA datasets are evaluated as follows.

\paragraph{VQA}
\begin{itemize}[noitemsep,nolistsep]
    \item \textbf{Accuracy} calculations follow \cite{vqa}: A question is 100\% correct if the returned answer was given by at least 3 (of 10) annotaters and otherwise contributes a fractional score ($min(\frac{\#annotaters}{3}, 1)$) to overall accuracy.
    \item \textbf{FPVG} for VQA-based tests requires customization of the original metric from \cite{reich2023fpvg}. We follow the process described in (and shared by) \cite{reich2024truevg}, where FPVG was customized for VQA-HAT.  \\
    Note that there is a mismatch between FPVG and accuracy numbers for VQA-based tests (e.g., in Table \ref{table:ood_tests}). In the original FPVG metric formulation, GGC+BGC equals accuracy. This is not the case in printed results for VQA-based tests, because accuracy for VQA-based datasets is calculated with fractional correctness scores (see above), while FPVG assigns binary values (i.e., the answer either counts as fully correct or wrong). Within FPVG, an answer counts as correct if the predicted answer is among the 10 reference answers, i.e., if at least one of the 10 annotaters gave this answer.
\end{itemize}

\paragraph{GQA}
\begin{itemize}[noitemsep,nolistsep]
    \item \textbf{Accuracy} in GQA-based tests is calculated as the number of correct answers divided by all questions.
    \item \textbf{FPVG} for GQA-based tests is determined as originally described in \cite{reich2023fpvg}.
\end{itemize}

\subsection{Model Training Details}
\label{appsec:model_training_details}
Our experiments make use of implementations shared by \cite{ying2022visfis} (UpDn, LXMERT), \cite{reich2024truevg} (symbolic features, Infusion, FPVG evaluation), \cite{lxmert} (LXMERT), \cite{mac} (MAC), \cite{chen2021meta} (MMN) and \cite{reich2023fpvg} (FPVG).

\subsubsection{UpDn}
UpDn models are all trained for 50 epochs on each evaluated dataset. Model selection after 50 epochs is based on performance on the held-out dev set. Other hyperparameters were adopted from code by \cite{reich2024truevg} (which builds on \cite{ying2022visfis}).

\subsubsection{LXMERT}
\textbf{Pre-training.} LXMERT \citep{lxmert} is pre-trained from scratch for 30 epochs for each dataset split individually (i.e., no mixing of datasets to avoid spoiling ID/OOD distributions). Model selection after 30 epochs is based on performance on the held-out dev set. We adjust the training scheme from the original paper as follows: 
\begin{itemize}[noitemsep,nolistsep]
    \item The attribute-related loss is only used for VQA-CP-based models, but not for GQA-based models (unsuitable for more than a single attribute per object, which GQA features provide).
    \item We use the original setting of 36 visual objects for VQA-CP and 100 objects for all GQA-based models.
    \item In all cases, we use a smaller version of the model to adjust to the reduced amount of training data of the examined data split: Hidden layer dimensions are reduced from 768 to 128. Intermediate layer size is reduced from 3072 to 512. Number of attention heads per self-attention layer is reduced from 12 to 4.
\end{itemize}
Pre-training is done individually for each dataset. For GQA-AUG experiments, we also apply INF training during pre-training, i.e., DET and INF evaluations of LXMERT involve individually pre-trained models (i.e., DET-pre-trained and INF-pre-trained).

\textbf{Fine-tuning.} We fine-tune each model for 35 epochs using a two-layered VQA classifier with softmax-based (GQA datasets) or sigmoid-based (VQA dataset) answer output. Model selection after 35 epochs is based on performance on the held-out dev set. Fine-tuning affects all LXMERT weights (not only the added VQA classifier). 

\subsubsection{MMN}
MMN \cite{chen2021meta} consists of two main modules that are trained separately: A program parser and the actual inference model, which takes the predicted program from the parser as input. We follow training recommendations from the official code-base with the following adjustments.
For the inference model, we run 12 epochs of ``bootstrapping'' with ground-truth programs and another 15 epochs of fine-tuning with parser-generated programs. In both cases, we use the GQA-AUG training set. Model selection is done in each run by early stopping of 1 epoch based on accuracy on the dev set. 
The program parser is the same across all instances.

\subsubsection{MAC}
MAC \cite{mac} is a monolithic multi-hop attention-based VQA model. We follow the official training procedure from the released code base. We train the model for 25 epochs and use early stopping based on performance on the dev set to select the best model.

\subsubsection{VLR}
VLR \cite{reich2022vlr} is a modular, symbolic method with rule-based (i.e., programmed) inference. Similar to MMN, it uses a program parser to generate a functional program from the input question in order to navigate the visual knowledge base (represented by a scene graph of the question-related image). The program parser is trained without involvement of visual features. The inference module has no trainable parameters. The program parser and inference module were built (and trained) according to the description in \cite{reich2022vlr}.

\section{Additional results}
\label{app:additional_results}

\subsection{OOD Tests}
\label{app:additional_results_ood}

We include additional results to accompany those for the four examined OOD splits from Table \ref{table:ood_tests}.
\begin{itemize}[noitemsep,nolistsep]
    \item \textbf{Max deviation:} Table \ref{apptable:ood_tests_idonly} (ID) and Table \ref{apptable:ood_tests_oodonly} (OOD) show the same results as Table \ref{table:ood_tests}, but in addition to the mean of measurements also include the max deviation of measurements over five differently seeded training runs for each model.
    \item \textbf{\textit{Query-}/\textit{other}-questions:} Based on the same models, we report numbers for \textit{query-}questions (GQA) and \textit{other-}questions (VQA). Test set sizes for these questions are listed in Table \ref{apptable:gqa_ood_dataset_query}. These questions are more challenging than binary-type questions in the dataset and less prone to be solved by SCs (see also discussion in \cite{ood_testing_teney}). \\ Table \ref{apptable:ood_tests_idonly_query} (ID) and Table \ref{apptable:ood_tests_oodonly_query} (OOD) show mean and max deviation of measurements over five differently seeded training runs for each model for these subsets.
\end{itemize}

\begin{table}[h]
\centering
\caption{Dataset sizes, tests only include \textit{other}-type (VQA) and \textit{query}-type questions (GQA).}
\label{apptable:gqa_ood_dataset_query}
\resizebox{0.5\textwidth}{!}{%
\begin{tabular}{lrrrr}
\toprule
Dataset & Train & Dev & ID & OOD \\
\midrule
GQA-CP-large \citep{ying2022visfis} & 645k & 107k & 73k & 74k \\
GQA-OOD \citep{kervadec2021roses} & 923k & 20k & 18k & 10k \\
VQA-CPv2 \citep{vqacp} & 418k & 20k & n/a & 8.7k \\
VQA-HAT-CP \citep{ying2022visfis} & 36k & 6k & 3.3k & 4.7k \\
\bottomrule
\end{tabular}
}  
\end{table}

\begin{table*}[h]
\centering
\caption{Mean ID accuracy and FPVG results and max deviation over five differently seeded training runs. Evaluating UpDn and LXMERT on four OOD test splits.} 
\resizebox{1\textwidth}{!}{%
\begin{tabular}{lcrrrrrrr}
\toprule
 \multicolumn{2}{c}{Training} & \multicolumn{7}{c}{ID} \\
 \cmidrule{3-9}
 Dataset & Model & Acc & $FPVG_{+}$ && GGC & GGW & BGC & BGW  \\
\midrule
 GQA-CP-large & UpDn & 64.53$\pm$0.15 & 24.30$\pm$0.43 && 18.65$\pm$0.26 & 5.65$\pm$0.20 & 45.89$\pm$0.13 & 29.82$\pm$0.30 \\
 & LXM & 70.05$\pm$0.08 & 24.13$\pm$0.25 && 20.16$\pm$0.16 & 3.97$\pm$0.12 & 49.89$\pm$0.19 & 25.98$\pm$0.17 \\
\midrule
GQA-OOD & UpDn &  63.18$\pm$0.75 & 27.52$\pm$0.60 && 20.95$\pm$0.32 & 6.58$\pm$0.33 & 42.23$\pm$0.44 & 30.25$\pm$0.99 \\
& LXM & 65.54$\pm$0.29 & 26.59$\pm$0.67 && 21.31$\pm$0.55 & 5.28$\pm$0.17 & 44.23$\pm$0.43 & 29.18$\pm$0.44 \\
\midrule
VQA-CPv2 & UpDn &  &  && n/a &  &  &    \\
 & LXM &  &  && n/a &  &  &   \\
\midrule
VQA-HAT-CP & UpDn & 54.49$\pm$0.14 & 23.14$\pm$2.44 && 12.89$\pm$1.73 & 10.25$\pm$0.89 & 46.52$\pm$1.97 & 30.34$\pm$0.73 \\
\bottomrule
\end{tabular}
} 
\label{apptable:ood_tests_idonly}
\end{table*}
\begin{table*}[h]
\centering
\caption{Mean OOD accuracy and FPVG results and max deviation over five differently seeded training runs. Evaluating UpDn and LXMERT on four OOD test splits.} 
\resizebox{1\textwidth}{!}{%
\begin{tabular}{lcrrrrrrr}
\toprule
 \multicolumn{2}{c}{Training} & \multicolumn{7}{c}{OOD} \\
 \cmidrule{3-9}
 Dataset & Model & Acc & $FPVG_{+}$ && GGC & GGW & BGC & BGW  \\
\midrule
 GQA-CP-large & UpDn & 44.60$\pm$0.66 & 23.46$\pm$0.46 && 14.94$\pm$0.28 & 8.52$\pm$0.36 & 29.66$\pm$0.46 & 46.87$\pm$0.85 \\
 & LXM & 53.51$\pm$0.20 & 23.69$\pm$0.47 && 17.29$\pm$0.35 & 6.40$\pm$0.14 & 36.22$\pm$0.35 & 40.09$\pm$0.29 \\
\midrule
GQA-OOD & UpDn & 43.72$\pm$0.62 & 26.78$\pm$0.64 && 15.82$\pm$0.17 & 10.97$\pm$0.47 & 27.90$\pm$0.47 & 45.32$\pm$0.60 \\
& LXM & 47.09$\pm$0.48 & 25.08$\pm$0.58 && 16.06$\pm$0.30 & 9.02$\pm$0.40 & 31.03$\pm$0.40 & 43.89$\pm$0.88 \\
\midrule
VQA-CPv2 & UpDn & 41.53$\pm$0.37 & 23.54$\pm$0.94 && 14.16$\pm$0.27 & 9.38$\pm$0.84 & 36.15$\pm$0.45 & 40.31$\pm$0.49 \\
 & LXM & 42.24$\pm$0.27 & 17.44$\pm$0.53 && 11.30$\pm$0.30 & 6.15$\pm$0.34 & 39.81$\pm$0.30 & 42.74$\pm$0.44  \\
\midrule
VQA-HAT-CP & UpDn & 40.80$\pm$1.56 & 26.57$\pm$2.50 && 11.83$\pm$1.86 & 14.74$\pm$0.94 & 33.94$\pm$0.81 & 39.49$\pm$2.07 \\
\bottomrule
\end{tabular}
} 
\label{apptable:ood_tests_oodonly}
\end{table*}
\begin{table*}[h]
\centering
\caption{Mean ID accuracy and FPVG results and max deviation over five differently seeded training runs. Evaluating UpDn and LXMERT on four OOD test splits. Tests only include \textit{query}-type questions (GQA) and \textit{other}-type questions (VQA).} 
\resizebox{1\textwidth}{!}{%
\begin{tabular}{lcrrrrrrr}
\toprule
 \multicolumn{2}{c}{Training} & \multicolumn{7}{c}{ID (\textit{query}/\textit{other}-type questions)} \\
 \cmidrule{3-9}
 Dataset & Model & Acc & $FPVG_{+}$ && GGC & GGW & BGC & BGW  \\
\midrule
 GQA-CP-large & UpDn & 54.31$\pm$0.17 & 30.88$\pm$0.13 && 23.00$\pm$0.13 & 7.87$\pm$0.23 & 31.31$\pm$0.10 & 37.82$\pm$0.10 \\
 & LXM & 59.54$\pm$0.24 & 30.36$\pm$0.41 && 24.37$\pm$0.22 & 5.99$\pm$0.19 & 35.18$\pm$0.26 & 34.46$\pm$0.32 \\
\midrule
GQA-OOD & UpDn & 54.03$\pm$0.52 & 32.03$\pm$0.78 && 23.47$\pm$0.34 & 8.56$\pm$0.46 & 30.56$\pm$0.56 & 37.41$\pm$0.96  \\
& LXM & 54.87$\pm$0.39 & 30.73$\pm$1.00 && 23.33$\pm$0.77 & 7.41$\pm$0.32 & 31.54$\pm$0.49 & 37.72$\pm$0.63 \\
\midrule
VQA-CPv2 & UpDn &  &  && n/a &  &  &    \\
 & LXM &  &  && n/a &  &  &   \\
\midrule
VQA-HAT-CP & UpDn & 41.18$\pm$0.49 & 32.89$\pm$2.38 && 16.73$\pm$1.17 & 16.16$\pm$1.42 & 28.51$\pm$1.27 & 38.61$\pm$1.70 \\
\bottomrule
\end{tabular}
} 
\label{apptable:ood_tests_idonly_query}
\end{table*}
\begin{table*}[h!]
\centering
\caption{Mean OOD accuracy and FPVG results and max deviation over five differently seeded training runs. Evaluating UpDn and LXMERT on four OOD test splits. Tests only include \textit{query}-type questions (GQA) and \textit{other}-type questions (VQA).} 
\resizebox{1\textwidth}{!}{%
\begin{tabular}{lcrrrrrrr}
\toprule
 \multicolumn{2}{c}{Training} & \multicolumn{7}{c}{OOD (\textit{query}/\textit{other}-type questions)} \\
 \cmidrule{3-9}
 Dataset & Model & Acc & $FPVG_{+}$ && GGC & GGW & BGC & BGW  \\
\midrule
 GQA-CP-large & UpDn & 34.93$\pm$0.43 & 28.74$\pm$0.27 && 16.17$\pm$0.23 & 12.57$\pm$0.45 & 18.76$\pm$0.59 & 52.50$\pm$0.79 \\
 & LXM & 41.09$\pm$0.19 & 28.04$\pm$0.29 && 17.90$\pm$0.14 & 10.14$\pm$0.20 & 23.20$\pm$0.29 & 48.77$\pm$0.36 \\
\midrule
GQA-OOD & UpDn & 31.14$\pm$0.39 & 28.97$\pm$0.72 && 15.17$\pm$0.16 & 13.79$\pm$0.61 & 15.96$\pm$0.21 & 55.07$\pm$0.77  \\
& LXM & 32.77$\pm$0.50 & 27.37$\pm$0.72 && 15.12$\pm$0.42 & 12.25$\pm$0.48 & 17.65$\pm$0.26 & 54.98$\pm$0.98 \\
\midrule
VQA-CPv2 & UpDn & 47.99$\pm$0.53 & 33.34$\pm$0.84 && 21.34$\pm$0.52 & 12.00$\pm$0.87 & 33.55$\pm$0.79 & 33.11$\pm$0.84 \\
 & LXM &  49.93$\pm$0.30 & 26.17$\pm$0.49 && 18.22$\pm$0.39 & 7.95$\pm$0.20 & 38.57$\pm$0.30 & 35.25$\pm$0.37 \\
\midrule
VQA-HAT-CP & UpDn & 31.71$\pm$2.23 & 32.53$\pm$2.68 && 14.07$\pm$1.96 & 18.46$\pm$1.68 & 22.07$\pm$0.78 & 45.41$\pm$2.95 \\
\bottomrule
\end{tabular}
} 
\label{apptable:ood_tests_oodonly_query}
\end{table*}

\subsection{GQA-AUG Experiments}
\label{app:additional_results_aug}
We provide additional details for Table \ref{table:INF_AUG_allmodels}, which only lists mean results over five differently seeded training runs, but not the max deviation. Table \ref{apptable:gqaaug_idonly} (ID) and Table \ref{apptable:gqaaug_oodonly} (OOD) list both mean and max deviation of results for all metrics and models.

\begin{table*}[h!]
\centering
\caption{GQA-AUG: Mean ID accuracy and FPVG results and max deviation over five differently seeded training runs.} 
\resizebox{1\textwidth}{!}{%
\begin{tabular}{lcrrrrrrr}
\toprule
 \multicolumn{2}{c}{Training} & \multicolumn{7}{c}{ID (GQA-AUG)} \\
 \cmidrule{3-9}
 Model & Infusion & Acc & $FPVG_{+}$ && GGC & GGW & BGC & BGW  \\
\midrule
 UpDn & no & 40.09$\pm$0.43 & 33.99$\pm$1.01 && 20.86$\pm$0.42 & 13.12$\pm$0.58 & 19.23$\pm$0.23 & 46.79$\pm$1.01 \\
 LXMERT & no & 41.95$\pm$0.28 & 30.78$\pm$0.94 && 19.50$\pm$0.66 & 11.28$\pm$0.57 & 22.44$\pm$0.66 & 46.77$\pm$0.84 \\
 MMN & no & 43.09$\pm$0.37 & 40.10$\pm$0.38 && 25.39$\pm$0.38 & 14.71$\pm$0.47 & 17.70$\pm$0.75 & 42.20$\pm$0.43 \\
 MAC & no & 40.44$\pm$0.48 & 31.64$\pm$0.42 && 19.76$\pm$0.37 & 11.88$\pm$0.30 & 20.68$\pm$0.78 & 47.68$\pm$0.51 \\
 \midrule
 VLR & n/a & 39.08 & 47.21 && 26.26 & 20.95 & 12.82 & 39.97\\
 \midrule
 UpDn & yes & 41.67$\pm$0.50 & 42.86$\pm$0.86 && 26.12$\pm$0.69 & 16.74$\pm$0.47 & 15.55$\pm$0.61 & 41.59$\pm$0.66 \\
 LXMERT & yes & 42.17$\pm$0.19 & 41.35$\pm$0.47 && 25.52$\pm$0.40 & 15.83$\pm$0.27 & 16.65$\pm$0.45 & 42.00$\pm$0.29 \\
 MMN & yes & 43.77$\pm$0.24 & 43.70$\pm$0.23 && 28.23$\pm$0.21 & 15.48$\pm$0.06 & 15.54$\pm$0.34 & 40.75$\pm$0.21 \\
 MAC & yes & 40.83$\pm$0.31 & 42.83$\pm$0.38 && 26.45$\pm$0.34  & 16.38$\pm$0.20 & 14.38$\pm$0.40 & 42.79$\pm$0.33 \\
\bottomrule
\end{tabular}
} 
\label{apptable:gqaaug_idonly}
\end{table*}
\begin{table*}[h!]
\centering
\caption{GQA-AUG: Mean OOD accuracy and FPVG results and max deviation over five differently seeded training runs.} 
\resizebox{1\textwidth}{!}{%
\begin{tabular}{lcrrrrrrr}
\toprule
 \multicolumn{2}{c}{Training} & \multicolumn{7}{c}{OOD (GQA-AUG)} \\
 \cmidrule{3-9}
 Model & Infusion & Acc & $FPVG_{+}$ && GGC & GGW & BGC & BGW  \\
\midrule
 UpDn & no & 16.27$\pm$1.52 & 27.27$\pm$1.14 && 12.57$\pm$1.36 & 14.70$\pm$1.12 & 3.70$\pm$0.29 & 69.03$\pm$1.14 \\
 LXMERT & no &  13.79$\pm$0.35 & 18.93$\pm$0.88 && 8.96$\pm$0.39 & 9.97$\pm$0.95 & 4.83$\pm$0.23 & 76.24$\pm$1.11 \\
 MMN & no & 21.07$\pm$0.57 & 27.01$\pm$0.16 && 16.94$\pm$0.30 & 10.07$\pm$0.39 & 4.13$\pm$0.72 & 68.86$\pm$0.86 \\
 MAC & no & 15.79$\pm$0.77 & 21.58$\pm$0.23 && 9.83$\pm$0.43 & 11.75$\pm$0.45 & 5.96$\pm$0.39 & 72.47$\pm$0.40 \\
 \midrule
 VLR & n/a & 81.23 & 78.76 && 77.26 & 1.50 & 3.97 & 17.27 \\
 \midrule
 UpDn & yes & 64.36$\pm$1.44 & 66.33$\pm$1.86 && 61.42$\pm$1.55 & 4.90$\pm$0.88 & 2.94$\pm$0.36 & 30.74$\pm$1.62 \\
 LXMERT & yes & 58.66$\pm$0.86 & 57.03$\pm$0.97 && 54.44$\pm$1.04 & 2.59$\pm$0.27 & 4.22$\pm$0.51 & 38.75$\pm$0.59 \\
 MMN & yes & 59.86$\pm$0.53 & 57.92$\pm$0.53 && 55.60$\pm$0.34 & 2.32$\pm$0.37 & 4.26$\pm$0.37 & 37.82$\pm$0.90 \\
 MAC & yes & 63.91$\pm$0.49 & 64.48$\pm$0.22 && 61.06$\pm$0.26 & 3.42$\pm$0.30 & 2.85$\pm$0.34 & 32.67$\pm$0.56 \\
\bottomrule
\end{tabular}
} 
\label{apptable:gqaaug_oodonly}
\end{table*}

\end{document}